\documentclass[11pt,a4paper]{article}
\usepackage[hyperref]{emnlp2018}
\usepackage{times}
\usepackage{latexsym}

\usepackage{balance}
\usepackage{url}

\usepackage{graphicx}
\usepackage{subfigure}
\usepackage{amssymb}
\usepackage{amsmath}
\usepackage{multirow}
\usepackage{makecell}
\usepackage{flushend}
\usepackage{color}
\usepackage{booktabs} 
\usepackage{xcolor}

\aclfinalcopy 

\newcommand*\samethanks[1][\value{footnote}]{\footnotemark[#1]}

\title{Deterministic Non-Autoregressive Neural Sequence Modeling \\ by Iterative Refinement}

\author{Jason Lee\thanks{~~~Equal Contribution} \\
  New York University \\
  {\tt jason@cs.nyu.edu} \\\And
  Elman Mansimov\samethanks \\
  New York University \\
  {\tt mansimov@cs.nyu.edu} \\\And
  Kyunghyun Cho \\
  New York University \\
  CIFAR Azrieli Global Scholar \\
  {\tt kyunghyun.cho@nyu.edu}
  }

\date{}

\begin{document}
\maketitle
\begin{abstract}
We propose a conditional non-autoregressive neural sequence model based on iterative refinement. The proposed model is designed based on the principles of latent variable models and denoising autoencoders, and is generally applicable to any sequence generation task. We extensively evaluate the proposed model on machine translation (En$\leftrightarrow$De and En$\leftrightarrow$Ro) and image caption generation, and observe that it significantly speeds up decoding while maintaining the generation quality comparable to the autoregressive counterpart.
\end{abstract}

\section{Introduction}

Conditional neural sequence modeling has become a {\it de facto} standard in a variety of tasks~\citep[see, e.g.,][and references therein]{cho2015describing}. Much of this recent success is built on top of autoregressive sequence models in which the probability of a target sequence is factorized as a product of conditional probabilities of next symbols given all the preceding ones. Despite its success, neural autoregressive modeling
has its weakness in decoding, i.e., finding the most likely sequence. Because of intractability, we must resort to suboptimal approximate decoding, and due to its sequential nature, decoding cannot be easily parallelized and results in a large latency~\citep[see, e.g.,][]{cho2016noisy}. This has motivated the recent investigation into non-autoregressive neural sequence modeling by \citet{gu2017non} in the context of machine translation and \citet{oord2017parallel} in the context of speech synthesis.

In this paper, we propose a non-autoregressive neural sequence model based on iterative refinement, which is generally applicable to any sequence generation task beyond machine translation. The proposed model can be viewed as both a latent variable model and a conditional denoising autoencoder. We thus propose a learning algorithm that is hybrid of 
lowerbound maximization
and 
reconstruction error minimization. We further design an iterative inference strategy with an adaptive number of steps to minimize the generation latency without sacrificing the generation quality.

We extensively evaluate the proposed conditional non-autoregressive sequence model and compare it against the autoregressive counterpart, using the state-of-the-art Transformer~\citep{vaswani2017attention}, on machine translation and image caption generation. In the case of machine translation, the proposed deterministic non-autoregressive models are able to decode approximately $2-3\times$ faster than beam search from the autoregressive counterparts on both GPU and CPU, while maintaining 90-95\% of translation quality on IWSLT'16 En$\leftrightarrow$De, WMT'16 En$\leftrightarrow$Ro and WMT'14 En$\leftrightarrow$De. On image caption generation, we observe approximately $3\times$ and $5\times$ faster decoding on GPU and CPU, respectively, while maintaining 85\% of caption quality.\footnote{
We release the implementation, preprocessed datasets as well as trained models online at \url{https://github.com/nyu-dl/dl4mt-nonauto}.
}

\section{Non-Autoregressive Sequence Models}

Sequence modeling in deep learning has largely focused on autoregressive modeling.
That is, given a sequence $Y=(y_1, \ldots, y_T)$, we use some form of a neural network to parametrize the conditional distribution over each variable $y_t$ given all the preceding variables, i.e.,
\[
\log p(y_t | y_{<t}) = f_{\theta}(y_{<t}),
\]
where $f_{\theta}$ is for instance a recurrent neural network. This approach has become a {\it de facto} standard in language modeling~\citep{mikolov2010recurrent}.
When this is 
conditioned on an extra variable
$X$, it becomes a conditional sequence model $\log p(Y|X)$ which serves as a basis on which many recent advances in, e.g., machine translation~\citep{bahdanau2014neural,sutskever2014sequence,kalchbrenner2013recurrent} and speech recognition~\citep{chorowski2015attention,chiu2017state} have been made. 

Despite the recent success, autoregressive sequence modeling has a weakness due to its nature of sequential processing. This weakness shows itself especially when we try to decode the most likely sequence from a trained model, i.e., 
\[
\hat{Y} = \arg\max_{Y} \log p(Y|X).
\] 
There is no known polynomial algorithm for solving it exactly, and practitioners have relied on approximate decoding algorithms
~\citep[see, e.g.,][]{cho2016noisy,hoang2017decoding}. Among these, beam search has become the method of choice, due to its superior performance over greedy decoding, which however comes with a substantial computational overhead~\citep{cho2016noisy}.

As a solution to this issue of slow decoding, two recent works have attempted non-autoregressive sequence modeling. \citet{gu2017non} have modified the 
Transformer~\citep{vaswani2017attention} for non-autoregressive machine translation, and \citet{oord2017parallel} a convolutional network~\citep{van2016wavenet} for non-autoregressive modeling of waveform. Non-autoregressive modeling 
factorizes
the distribution over a target sequence given a source into a product of conditionally independent per-step distributions:
\[
p(Y|X) = \prod_{t=1}^T p(y_t | X),
\]
breaking the dependency among the target variables across time. 
This 
allows us to trivially find the most likely target sequence by taking \mbox{$\arg\max_{y_t} p(y_t | X)$} for each $t$, effectively bypassing the computational overhead and sub-optimality of decoding from an autoregressive sequence model. 

This desirable property of exact and parallel decoding however comes at the expense of potential performance degradation~\citep{kaiser2016can}. The potential modeling gap, which is the gap between the underlying, true model and the neural sequence model, could be larger with the non-autogressive model compared to the autoregressive one due to challenge of modeling the factorized conditional distribution above.


\section{Iterative Refinement for Deterministic Non-Autoregressive Sequence Models}

\subsection{Latent variable model}

Similarly to two recent works~\citep{oord2017parallel,gu2017non}, we introduce latent variables to implicitly capture the dependencies among target variables. We however remove any stochastic behavior by interpreting this latent variable model, introduced immediately below, as a process of iterative refinement. 

Our goal is to capture the dependencies among target symbols 
given a source sentence 
without auto-regression
by introducing $L$
intermediate random variables and marginalizing them out:
\begin{align}
\label{eq:original}
&p(Y|X) 
= \sum_{Y^{0}, \ldots, Y^{L}}  
\left(\prod_{t=1}^T p(y_t|Y^{L}, X)\right)\\
&\left(\prod_{t=1}^T p(y^{L}_t|Y^{L-1}, X)\right) 
\cdots 
\left(\prod_{t=1}^T p(y^{0}_t | X)\right).
\nonumber
\end{align}
Each product term inside the summation is modelled by a deep neural network that takes as input a source sentence
and outputs the conditional distribution over the target vocabulary $V$ for each $t$. 

\paragraph{Deterministic Approximation} 

The marginalization in Eq.~\eqref{eq:original} is intractable. In order to avoid this issue, we consider two approximation strategies;  deterministic and stochastic approximation. Without loss of generality, let us consider the case of single intermediate latent variable, that is $L=1$. In the {\it deterministic} case, we set $\hat{y}^{0}_t$ to the most likely value according to its distribution $p(y^{0}_t| X)$, that is $\hat{y}^{0}_t = \arg\max_{y^{0}_t} p(y^{0}_t |X)$. The entire lower bound can then be written as:
\begin{align*}
&\log p(Y|X) 
\geq  \left(\sum_{t=1}^T \log p(y_t|\hat{Y}^{L}, X)\right) + \cdots 
\nonumber
\\
& + \left(\sum_{t=1}^T \log p(y^1_t|\hat{Y}^{0}, X)\right) +
\left(\sum_{t=1}^T \log p(\hat{y}^{0}_t | X)\right).
\end{align*}

\paragraph{Stochastic Approximation}

In the case of {\it stochastic} approximation, we instead sample $\hat{y}^{0}_t$ from the distribution $p(y_t^0|X)$. This results in the unbiased estimate of the marginal log-probability $\log p(Y|X)$. Other than the difference in whether most likely values or samples are used, the remaining steps are identical.


\paragraph{Latent Variables} 

Although the intermediate random variables could be anonymous, we constrain them to be of the same type as the output $Y$ is, in order to share an underlying neural network. This constraint allows us to view each conditional $p(Y^l|\hat{Y}^{l-1}, X)$ as a single-step of refinement of a rough target sequence $\hat{Y}^{l-1}$. The entire chain of $L$ conditionals is then the $L$-step iterative refinement. Furthermore, sharing the parameters across these refinement steps enables us to dynamically adapt the number of iterations per input $X$. This is important as it substantially reduces the amount of time required for decoding, as we see later in the experiments.

\paragraph{Training}

For each training pair $(X,Y^*)$, we first approximate the marginal log-probability.
We then minimize 
\begin{align}
\label{eq:lvm}
J_{\text{LVM}}(\theta) = -\sum_{l=0}^{L+1} \left(\sum_{t=1}^T \log p_{\theta} (y^*_t|\hat{Y}^{l-1}, X)\right),
\end{align}
where $\hat{Y}^{l-1}=(\hat{y}_1^{l-1}, \ldots, \hat{y}_T^{l-1})$, and $\theta$ is a set of parameters. We initialize $\hat{y}^0_{t}$ ($t$-th target word in the first iteration) as $x_{t'}$, where $t'=(T'/T)\cdot t$. $T'$ and $T$ are the lengths of the source $X$ and target $Y^*$, respectively. 

\subsection{Denoising Autoencoder}
\label{sec:dae}

The proposed approach could instead be viewed as learning a conditional denoising autoencoder which is known to capture the gradient of the log-density.
That is, we implicitly learn to find a direction $\Delta_Y$ in the output space that maximizes the underlying true, data-generating distribution $\log P(Y|X)$. Because the output space is discrete, much of the theoretical analysis by \citet{alain2014regularized} are not strictly applicable. We however find this view attractive as it serves as an alternative foundation for designing a learning algorithm.

\paragraph{Training} 

We start with a corruption process $C(Y|Y^*)$, which introduces noise to the correct output $Y^*$. Given 
the reference translation $Y^*$, we sample $\tilde{Y} \sim C(Y|Y^*)$ 
which becomes as an input to each conditional in Eq.~\eqref{eq:original}. Then, the goal of learning is to maximize the log-probability of the original reference $Y^*$ given the corrupted version. That is, to minimize

\begin{align}
\label{eq:denoising}
J_{\text{DAE}}(\theta) = -\sum_{t=1}^T \log p_{\theta} (y^*_t|\tilde{Y}, X).
\end{align}

Once this cost $J_{\text{DAE}}$ is minimized, we can recursively perform the maximum-a-posterior inference, i.e., 
\mbox{$\hat{Y} = \arg\max_{Y} \log p_{\theta}(Y|X),$}
to find $\hat{Y}$ that (approximately) maximizes $\log p(Y|X)$.

\paragraph{Corruption Process $C$}

There is little consensus on the best corruption process for a sequence, especially of discrete tokens. In this work, we use a corruption process proposed by \citet{Hill:16}, which has recently become more widely adopted~\citep[see, e.g.,][]{Artetxe:17,Lample:17}. Each $y^*_t$ in a reference target \mbox{$Y^*=(y_1^*, \ldots, y^*_{T})$} is corrupted with a probability \mbox{$\beta \in \left[0, 1\right]$}. If decided to corrupt, we either (1) replace $y^*_{t+1}$ with this token $y^*_t$, (2) replace $y^*_t$ with a token uniformly selected from a vocabulary of all unique tokens at random, or (3) swap $y^*_{t}$ and $y^*_{t+1}$. This is done sequentially from $y^*_1$ until $y^*_{T}$.

\subsection{Learning}

\paragraph{Cost function}

Although it is possible to train the proposed non-autoregressive sequence model using either of the cost functions above ($J_{\text{LVM}}$ or $J_{\text{DAE}}$,) we propose to stochastically mix these two cost functions. We do so by randomly replacing each term $\hat{Y}^{l-1}$ in Eq.~\eqref{eq:lvm} with $\tilde{Y}$ in Eq.~\eqref{eq:denoising}:
\begin{align}
\label{eq:final_cost}
J(\theta) = -\sum_{l=0}^{L+1} &\left(
\alpha_l \sum_{t=1}^T \log p_{\theta} (y^*_t|\hat{Y}^{l-1}, X)
\right. 
\\
&\left.
+(1-\alpha_l) \sum_{t=1}^T \log p_{\theta} (y^*_t|\tilde{Y}, X)
\right),
\nonumber
\end{align}
where $\tilde{Y} \sim C(Y|Y^*)$, and $\alpha_l$ is a sample from a Bernoulli distribution with the probability $p_{\text{DAE}}$.
$p_{\text{DAE}}$ is a hyperparameter. As the first conditional $p(Y^0 | X)$ in Eq.~\eqref{eq:original} does not take as input any target $Y$, we set $\alpha_0=1$ always.

\paragraph{Distillation}

\citet{gu2017non}, in the context of machine translation, and \citet{oord2017parallel}, in the context of speech generation, have recently discovered that it is important to use knowledge distillation~\citep{hinton2015distilling,kim2016sequence} to successfully train a non-autoregressive sequence model. Following \citet{gu2017non}, we also use knowledge distillation by replacing the reference target $Y^*$ of each training example $(X,Y^*)$ with a target $Y^{\text{AR}}$ generated from a well-trained autoregressive counterpart. Other than this replacement, the cost function in Eq~\eqref{eq:final_cost} and the model architecture remain unchanged. 

\paragraph{Target Length Prediction}

One difference between the autoregressive and non-autoregressive models is that the former naturally models the length of a target sequence without any arbitrary upper-bound, while the latter does not. It is hence necessary to separately model $p(T|X)$, where $T$ is the length of a target sequence, although during training, we simply use the length of each reference target sequence. 

\subsection{Inference: Decoding}
\label{sec:inference}

Inference in the proposed approach is entirely deterministic. We start from the input $X$ and first predict the length of the target sequence \mbox{$\hat{T} = \arg\max_{T} \log p(T|X)$}. Then, given $X$ and $\hat{T}$ we generate the initial target sequence by 
\mbox{$\hat{y}_t^0 = \arg\max_{y_t} \log p(y_t^0 | X)$},
for \mbox{$t=1,\ldots, T$}
We continue refining the target sequence 
by
\mbox{$\hat{y}_t^l = \arg\max_{y_t} \log p(y_t^l | \hat{Y}^{l-1}, X)$},  for  \mbox{$t=1,\ldots, T$}.

Because these conditionals, except for the initial one, are modeled by a single, shared neural network, this refinement can be performed as many iterations as necessary
until a predefined stopping criterion is met. A criterion can
be based either on the amount of change in a target sequence after each iteration (i.e., \mbox{$D(\hat{Y}^{l-1}, \hat{Y}^{l}) \leq \epsilon$}), or on the amount of change in the conditional log-probabilities (i.e., \mbox{$|\log p(\hat{Y}^{l-1}|X) - \log p(\hat{Y}^{l-1}|X)|\leq \epsilon$}) or on the computational budget. In our experiments, we use the first criterion and use Jaccard distance as our distance function $D$.

\section{Related Work}

\paragraph{Non-Autoregressive Neural Machine Translation} 

\citet{schwenk2012continuous} proposed a continuous-space translation model
to estimate the conditional distribution over a target phrase given a source phrase, while dropping the conditional dependencies among target tokens.
The evaluation was however limited to reranking and to short phrase pairs (up to 7 words on each side) only. 
\citet{kaiser2016can} investigated 
neural GPU~\citep{kaiser2015neural}, for machine translation. They evaluated both non-autoregressive and autoregressive approaches, and found that the non-autoregressive approach significantly lags behind the autoregressive variants. 
It however differs from our approach that each iteration does not output a refined version from the previous iteration. 
The recent paper by \citet{gu2017non} is most relevant to the proposed work. 
They similarly introduced a sequence of discrete latent variables. 
They however use supervised learning for inference,
using the
word alignment tool~\citep{dyer2013simple}.
To achieve the best result, \citet{gu2017non} stochastically sample the latent variables and rerank the corresponding target sequences with an external, autoregressive model. This is in 
contrast to the proposed approach which is fully deterministic during decoding and does not rely on any extra reranking mechanism. 

\begin{figure*}[t!]
    \centering
    \begin{minipage}{0.35\textwidth}
        \centering
        \includegraphics[width=1.0\columnwidth]{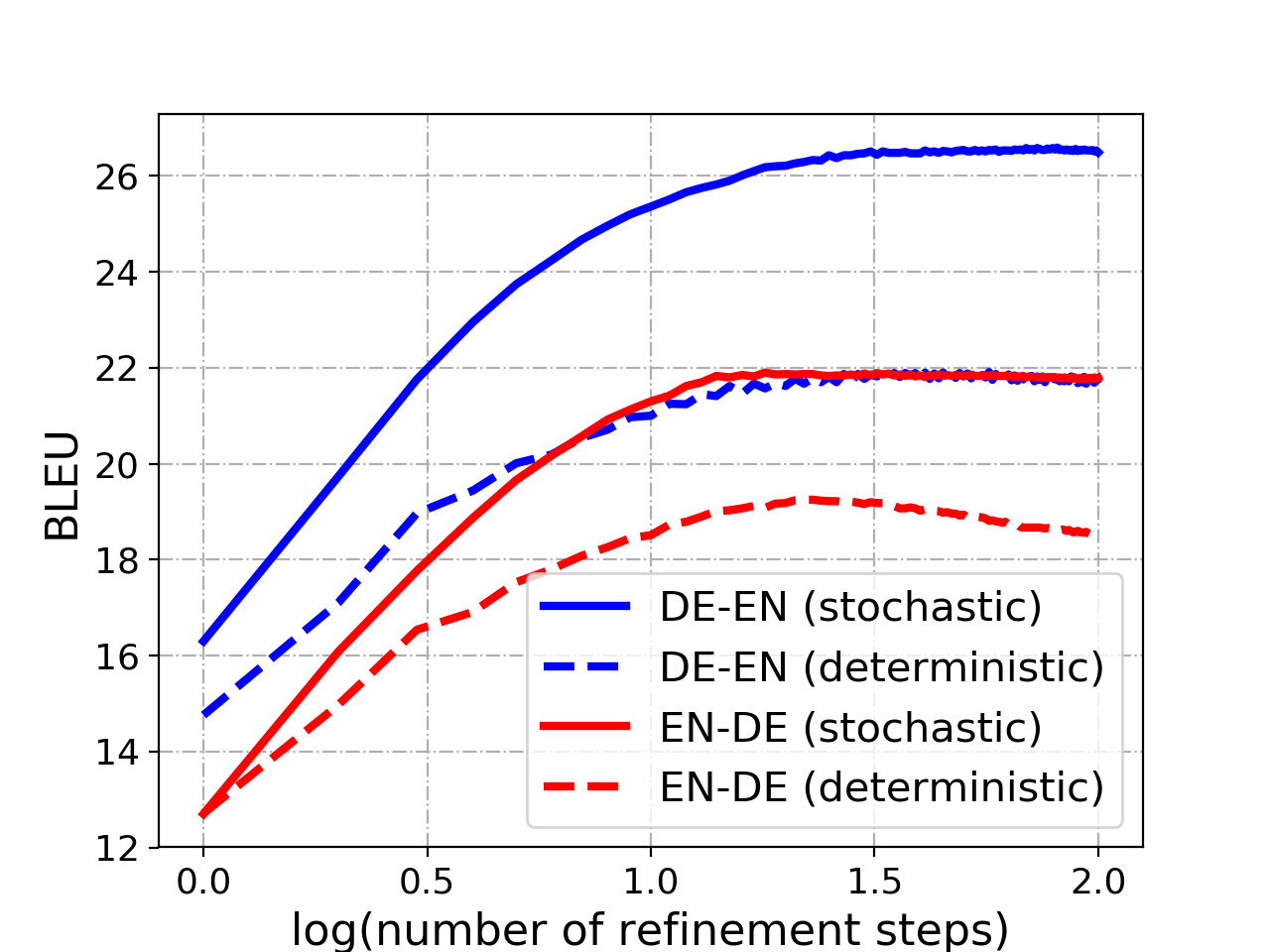} 
        (a)
\end{minipage}
    \hfill
    \begin{minipage}{0.35\textwidth}
        \centering
        \includegraphics[width=1.0\columnwidth]{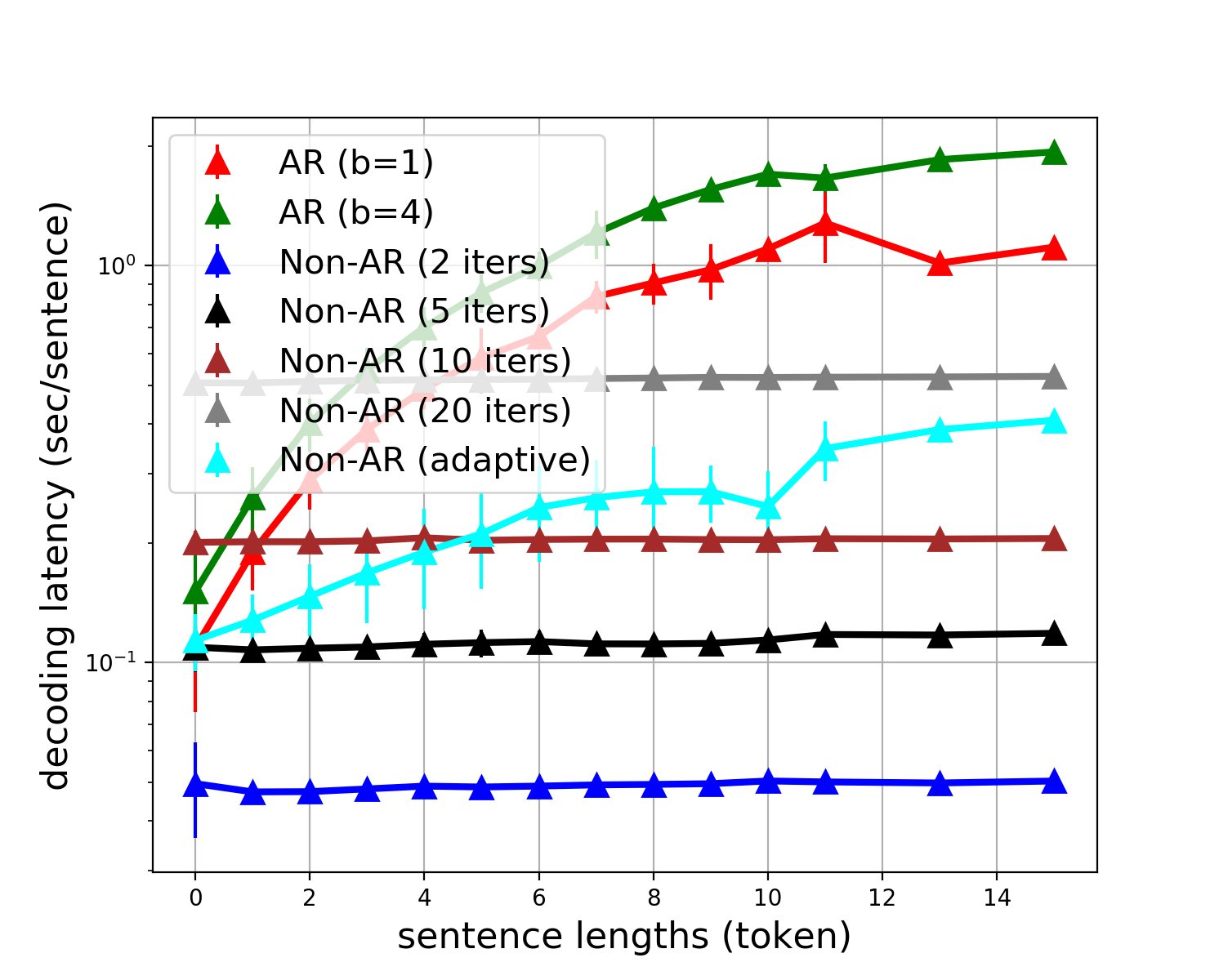} 
        (b)
    \end{minipage}
    \hfill
    \begin{minipage}{0.28\textwidth}
        \caption{
        (a) BLEU scores on WMT'14 En-De w.r.t. the number of refinement steps (up to $10^2$). The x-axis is in the logarithmic scale.
        (b) the decoding latencies (sec/sentence) of different approaches on IWSLT'16 En$\rightarrow$De. The y-axis is in the logarithmic scale.}
        \label{fig:latency}
        %
    \end{minipage}
    \vspace{-6mm}
\end{figure*}

\paragraph{Parallel WaveNet} 

Simultaneously with \citet{gu2017non}, \citet{oord2017parallel} presented a 
non-autoregressive sequence model for speech generation. They use inverse autoregressive flow~\citep[IAF,][]{kingma2016improved} to map a sequence of independent random variables to a target sequence. 
They 
apply the IAF multiple times, similarly to our iterative refinement strategy. Their approach is however restricted to continuous target variables, while the proposed approach in principle could be applied to both discrete and continuous variables.

\paragraph{Post-Editing for Machine Translation}

\citet{auli_iterative_mt} proposed a convolutional neural network that iteratively predicts and applies token substitutions given a translation from a phase-based translation system. Unlike their system, our approach can edit an intermediate translation with a higher degree of freedom.
QuickEdit~\citep{quickedit} and deliberation network~\citep{xia2017deliberation} incorporate the idea of refinement into neural machine translation. Both systems consist of two autoregressive decoders. The second decoder takes into account the translation generated by the first decoder. 
We 
extend these earlier efforts by incorporating more than one refinement steps without necessitating extra annotations. 

\paragraph{Infusion Training}

\citet{pascal_infusion} proposed an unconditional generative model for images based on iterative refinement.
At each step $l$ of iterative refinement, the model is trained to maximize the log-likelihood of target $Y$ given the weighted mixture of generated samples from the previous iteration $\hat{Y}^{l-1}$ and a corrupted target $\tilde{Y}$. That is, the corrupted version of target is ``infused'' into generated samples during training. In the domain of text, however, computing a weighted mixture of two sequences of discrete tokens is not well defined, and we propose to stochastically mix denoising and lowerbound maximization objectives. 


\section{Network Architecture}
\label{sec:architecture}

\begin{figure}[t]
\centering
\includegraphics[width=\columnwidth]{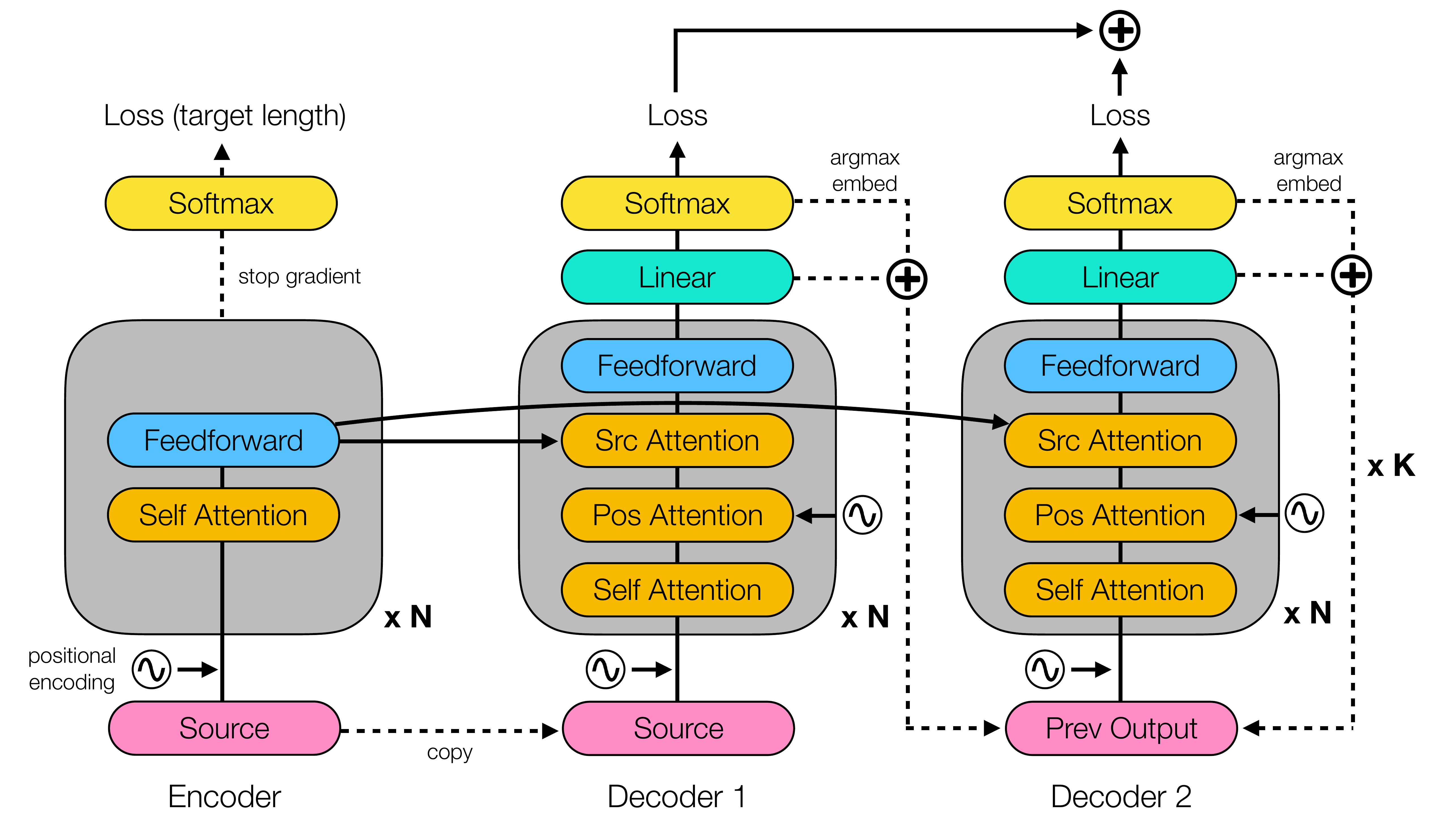}

\vspace{-3mm}
\caption{We compose three transformer blocks (``Encoder'', ``Decoder 1'' and ``Decoder 2'') to implement the proposed non-autoregressive sequence model. 
}
\label{fig:model}
\vspace{-5mm}
\end{figure}

We use three transformer-based network blocks to implement our model. 
The first block (``Encoder'') encodes the input $X$, the second block (``Decoder 1'') models the first conditional $\log p(Y^0|X)$, and the final block (``Decoder 2'') is shared across iterative refinement steps, modeling $\log p(Y^l|\hat{Y}^{l-1}, X)$. These blocks are depicted side-by-side in Fig.~\ref{fig:model}. The encoder is identical to that from the original Transformer~\citep{vaswani2017attention}. We however use the decoders from \citet{gu2017non} with additional positional attention and use the highway layer~\citep{srivastava2015highway} instead of the residual layer~\citep{he2016deep}.

The original input $X$ is padded or shortned to fit the length of the  reference target sequence before being fed to
Decoder 1. At each refinement step $l$, Decoder 2 takes as input the predicted target sequence $\hat{Y}^{l-1}$ and the sequence of final activation vectors from the previous step. 

{
\centering
\begin{table*}[t]
	\small
	\begin{center}
    \begin{tabular}[b]{llrrrr|rrrr|rrrr|rrr}
    \toprule
    & & \multicolumn{4}{c|}{IWSLT'16 En-De} & \multicolumn{4}{c|}{WMT'16 En-Ro} & \multicolumn{4}{c|}{WMT'14 En-De} & \multicolumn{3}{c}{MS COCO} \\
    & & En$\rightarrow$ & De$\rightarrow$ & GPU & CPU & En$\rightarrow$ & Ro$\rightarrow$ & GPU & CPU & En$\rightarrow$ & De$\rightarrow$ & GPU & CPU & BLEU & GPU & CPU \\
    
   	\midrule
     \multirow{2}{*}{\rotatebox[origin=c]{90}{\scriptsize AR}}
		& $b$\;=\;1 			& 28.64 & 34.11 &  70.3 &   32.2 & 31.93 & 31.55 &  55.6 &  15.7 & 23.77 & 28.15 &  54.0 &  15.8 & 23.47 &  4.3 & 2.1 \\
        & $b$\;=\;4 			& 28.98 & 34.81 &  63.8 &   14.6 & 32.40 & 32.06 &  43.3 &   7.3 & 24.57 & 28.47 &  44.9 & 7.0 & 24.78 &  3.6 & 1.0 \\
    \midrule
    \multirow{2}{*}{\rotatebox[origin=c]{90}{\scriptsize NAT}}
        & FT & 26.52 & -- & -- & -- & 27.29 & 29.06 & -- & -- & 17.69 & 21.47 & -- & -- & -- & -- & -- \\
        & FT+NPD & 28.16 & -- & -- & -- & 29.79 & 31.44 & -- & -- & 19.17 & 23.30 & -- & -- & -- & -- & -- \\  
    \midrule
     \multirow{6}{*}{\rotatebox[origin=c]{90}{\scriptsize Our Model}}
     & $i_\text{dec}$\;=\;1 	& 22.20  & 27.68 & 573.0 & 213.2 & 24.45 & 25.73 & 694.2 &  98.6 & 13.91 & 16.77 & 511.4 & 83.3 & 20.12 & 17.1 & 8.9 \\
     & $i_\text{dec}$\;=\;2 	& 24.82  & 30.23 & 423.8 & 110.9 & 27.10 & 28.15 & 332.7 &  62.8 & 16.95 & 20.39 & 393.6 & 49.6 & 20.88 & 12.0 & 5.7 \\
     & $i_\text{dec}$\;=\;5 	& 26.58  & 31.85 & 189.7 &  52.8 & 28.86 & 29.72 & 194.4 &  29.0 & 20.26 & 23.86 & 139.7 & 23.1 & 21.12 &  6.2 & 2.8 \\
     & $i_\text{dec}$\;=\;10 	& 27.11  & 32.31 &  98.8 &  24.1 & 29.32 & 30.19 &  93.1 &  14.8 & 21.61 & 25.48 & 90.4 & 12.3 & 21.24 &  2.0 & 1.2 \\
     \cmidrule{2-17}
     & Adaptive 			    & 27.01  & 32.43 & 125.9 &  29.3 & 29.66 & 30.30 & 118.3 &  16.5 & 21.54 & 25.43 & 107.2 &  20.3 & 21.12 & 10.8 & 4.8 \\
    \bottomrule
	\end{tabular}
    \vspace{-2mm}
	\caption{Generation quality (BLEU$\uparrow$) and decoding efficiency (tokens/sec$\uparrow$ for translation, images/sec$\uparrow$ for image captioning). Decoding efficiency is measured sentence-by-sentence.
	AR: autoregressive models. $b$: beam width. $i_{\text{dec}}$: the number of refinement steps taken during decoding. Adaptive: the adaptive number of refinement steps. NAT: non-autoregressive transformer models~\citep{gu2017non}. FT: fertility. NPD reranking using $100$ samples.
	}
	\label{tab:bleu_performance2}
	\end{center}
    \vspace{-6mm}
\end{table*}    
}

\section{Experimental Setting}

We evaluate the proposed approach on two 
sequence modeling tasks: 
machine translation and 
image caption generation. We compare the proposed non-autoregressive model against the autoregressive counterpart both in terms of generation quality, measured in terms of BLEU~\citep{papineni2002bleu}, and generation efficiency, measured in terms of (source) tokens and images per second for translation and image captioning, respectively.

\paragraph{Machine Translation} 

We choose three tasks 
of different sizes: IWSLT'16 En$\leftrightarrow$De (196k pairs), WMT'16 En$\leftrightarrow$Ro (610k pairs) and WMT'14 En$\leftrightarrow$De (4.5M pairs).
We tokenize each sentence
using a script from Moses~\citep{koehn2007moses} and segment each word into subword units using BPE~\citep{Sennrich:16}. We use 40k tokens from both source and target for all the tasks. For WMT'14 En-De, we use newstest-2013 and newstest-2014 as development and test sets. For WMT'16 En-Ro, we use newsdev-2016 and newstest-2016 as development and test sets. For IWSLT'16 En-De, we use test2013 for validation.

We closely follow the setting by \citet{gu2017non}. In the case of IWSLT'16 En-De, we use the small model ($d_{\text{model}}=278, d_\text{hidden}=507, p_\text{dropout}=0.1, n_{\text{layer}}=5$ and $n_\text{head}=2$).\footnote{
Due to the space constraint, we refer readers to \citep{vaswani2017attention,gu2017non} for more details.
} 
For WMT'14 En-De and WMT'16 En-Ro, we use the base transformer by \citet{vaswani2017attention} ($d_{\text{model}}=512, d_\text{hidden}=512, p_\text{dropout}=0.1, n_{\text{layer}}=6$ and $n_\text{head}=8$). We use the warm-up learning rate scheduling~\citep{vaswani2017attention} for the WMT tasks, while using linear annealing (from $3\times 10^{-4}$ to $10^{-5}$) for the IWSLT task.
We do not use label smoothing nor average multiple check-pointed models. These decisions were made based on the preliminary experiments. We train each model either on a single P40 (WMT'14 En-De and WMT'16 En-Ro) or on a single P100 (IWSLT'16 En-De) with each minibatch consisting of approximately 2k tokens. We use four P100's to train non-autoregressive models on WMT'14 En-De. 

\paragraph{Image Caption Generation: MS COCO}

We use MS COCO~\citep{mscoco}.
We use the publicly available splits 
~\citep{karpathy_caption}, consisting of 113,287 training images, 5k validation images and 5k test images. We extract
49 512-dimensional feature vectors for each image, using a ResNet-18~\citep{he2016deep} pretrained on ImageNet~\citep{deng2009imagenet}. The average of these 
vectors is copied as many times to match the length of the target sentence (reference during training and predicted during evaluation) to form the initial input to Decoder 1. We use the base transformer~\citep{vaswani2017attention} except that $n_{\text{layer}}$ is set to 4.
We train each model on a single 
1080ti with each minibatch consisting of approximately 1,024 tokens.

\paragraph{Target Length Prediction}

We formulate the target length prediction as classification, predicting the difference between the target and source lengths for translation and the target length for image captioning. All the hidden vectors from the $n_{\text{layer}}$ layers of the encoder are summed and fed to a softmax classifier after affine transformation. We however do not tune the encoder's parameters for target length prediction. We use this length predictor only during test time.
We find it important to accurately predict the target length for good overall performance. See Appendix~\ref{sec:lenpred} for an analysis on our length prediction model.

\paragraph{Training and Inference}

We use Adam~\citep{kingma2014adam}
and use $L=3$ in Eq.~\eqref{eq:original} during training
($i_{\text{train}}=4$ from hereon.) We use $p_{\text{DAE}}=0.5$. 
We use the deterministic strategy for IWSLT'16 En-De, WMT'16 En-Ro and MS COCO, while the stochastic strategy is used for WMT'14 En-De. These decisions were made based on the validation set performance.
After both the 
non-autogressive sequence model and target length predictor are trained, we decode by first predicting the target length and then running iterative refinement steps until the outputs of consecutive iterations are the same (or Jaccard distance between consecutive decoded sequences is $1$). To assess the effectiveness of this adaptive scheme, we also test a fixed number of steps ($i_{\text{dec}}$). 
In 
machine translation, we remove any repetition by collapsing multiple consecutive occurrences of a token.

\section{Results and Analysis}


We make some important observations 
in Table~\ref{tab:bleu_performance2}. First, the generation quality improves across all the tasks as we run more refinement steps $i_{\text{dec}}$ even beyond that used in training ($i_{\text{train}}=4$), which supports our interpretation
as a conditional denoising autoencoder in Sec.~\ref{sec:dae}. To further verify this, we run decoding on WMT'14 (both directions) up to 100 iterations. As shown in Fig.~\ref{fig:latency}~(a), the quality improves well beyond the number of refinement steps used during training.

Second, the generation efficiency decreases as more refinements are made. 
We plot the average seconds per sentence in Fig.~\ref{fig:latency}~(b), measured on GPU while sequentially decoding one sentence at a time. As expected, decoding from the autoregressive model linearly slows down as the sentence length 
grows, while decoding from the 
non-autoregressive model with a fixed number of iterations has the constant complexity. However, the generation efficiency of non-autoregressive model decreases as more refinements are made. To make a smooth trade-off between the quality and speed, the adaptive decoding scheme 
allows us to achieve near-best generation quality with a significantly lower computational overhead. Moreover, 
the adaptive decoding scheme automatically increases the number of refinement steps as the sentence length 
increases, suggesting that this scheme captures the amount of information in the input well. The increase in latency is however less severe than
that of the autoregressive model. 

We also observe that the speedup in decoding 
is much clearer on GPU than on CPU. This is a consequence of highly parallel computation of the proposed non-autoregressive model, which is better suited to GPUs, showcasing the potential of using the non-autoregressive model with a specialized hardware for parallel computation, such as Google's TPUs~\citep{jouppi2017datacenter}. 
The results of our model decoded with adaptive decoding scheme are comparable to the results from
\citep{gu2017non}, 
without relying on any external tool.
On WMT'14 En-De, the proposed model outperforms 
the best model from \citep{gu2017non} by two 
points.

Lastly, it is encouraging to observe that the proposed non-autoregressive model works well on image caption generation. This result confirms the generality of our approach beyond machine translation, unlike that by \citet{gu2017non} which was 
for machine translation or by \citet{oord2017parallel} which was for speech synthesis.

{
\centering
\begin{table}[t]
	\begin{center}
	\small
    \begin{tabular}[b]{lccc|rr|rr}
    \toprule
    & & & & \multicolumn{2}{c|}{En$\rightarrow$De} & \multicolumn{2}{c}{De$\rightarrow$En} \\
    &    $i_{\text{train}}$ & $p_{\text{DAE}}$ & distill 
    & rep &  no rep & rep & no rep \\
	\midrule
    \multirow{2}{*}{\rotatebox[origin=c]{90}{\footnotesize AR}}
    	& \multicolumn{3}{c|}{$b=1$} & \multicolumn{2}{c|}{ 28.64 } & \multicolumn{2}{c}{ 34.11 } \\
    	& \multicolumn{3}{c|}{$b=4$} & \multicolumn{2}{c|}{ 28.98 } & \multicolumn{2}{c}{ 34.81 } \\
    \midrule
    \multirow{6}{*}{\rotatebox[origin=c]{90}{\footnotesize Our Models}} 
    & $1$ & 0 & & 14.62 & 18.03 & 16.70 & 21.18  \\
    & $2$ & 0 & & 17.42 & 21.08 & 19.84 &24.25  \\
    & $4$ & 0 & & 19.22 & 22.65 & 22.15 & 25.24 \\
    & $4$ & 1 & & 19.83 & 22.29 & 24.00 & 26.57  \\
    & $4$ & 0.5 & & 20.91 & 23.65 & 24.05 & 28.18  \\
    & $4$ & 0.5 & $\surd$ & 26.17 & 27.11 & 31.92 & 32.59 \\    
    \bottomrule
	\end{tabular}
    \vspace{-4mm}
	\caption{Ablation study on the dev set of IWSLT'16. 
	}
	\label{tab:ablation}
	\end{center}
    \vspace{-7mm}
\end{table}
}

\paragraph{Ablation Study}

We 
use IWSLT'16 En-De 
to investigate the impact of different number of refinement steps during training (denoted as $i_{\text{train}}$) as well as probability of using denoising autoencoder objective during training (denoted as $p_{\text{DAE}}$). The results are presented in Table~\ref{tab:ablation}. 

First, we observe that it is beneficial to use multiple iterations of refinement during training. By using four iterations (one step of decoder 1, followed by three steps of decoder 2), the BLEU score improved by approximately 1.5 points in both directions. We also notice that it is necessary to use the proposed hybrid learning strategy
to maximize the improvement from more iterations during training ($i_{\text{train}}=4$ vs. $i_{\text{train}}=4,p_{\text{DAE}}=1.0$ vs. $i_{\text{train}}=4,p_{\text{DAE}}=0.5$.) Knowledge distillation was 
crucial to close the gap between the proposed deterministic non-autoregressive sequence model and its autoregressive counterpart, echoing the observations by \citet{gu2017non} and \citet{oord2017parallel}. Finally, 
we see that removing repeating consecutive symbols
improves the quality of the best trained models ($i_{\text{train}}=4,p_{\text{DAE}}=0.5$) by approximately +1 BLEU. This suggests that the proposed iterative refinement is not enough to remove repetitions on its own. Further investigation 
is necessary to properly tackle this issue, which we leave as a future work.

{
\centering
\begin{table}[t]
    \begin{center}
    \small
    \begin{tabular}[b]{cc|rr}
    \toprule
    stochastic & distill & IWSLT'16 (En$\rightarrow$) & WMT'14 (En$\rightarrow$) \\
    \midrule
    & & 23.65 & 7.56 \\
    $\surd$ & & 22.80 & 16.56 \\
    \midrule
    & $\surd$ & 27.11 & 18.91 \\
    $\surd$ & $\surd$ & 25.39 & 21.22 \\
    \bottomrule
	\end{tabular}
    \vspace{-4mm}
	\caption{
	Deterministic and stochastic approximation}
	\label{tab:ablation_lowerbound}
	\end{center}
    \vspace{-7mm}
\end{table}
}

We then compare the deterministic and stochastic approximation strategies on IWSLT'16 En$\rightarrow$De and WMT'14 En$\rightarrow$De.
According to the results in Table~\ref{tab:ablation_lowerbound}, the stochastic strategy is crucial with a large corpus (WMT'14), while the deterministic strategy works as well or better with a small corpus (IWSLT'16). Both of the strategies benefit from knowledge distillation, but the gap between the two strategies when the dataset is large is much more apparent without knowledge distillation.





\subsection{Qualitative Analysis}

{
\centering
\begin{table*}[t]
\begin{minipage}{\textwidth}
\small
\begin{tabular}{ll}
    \midrule
    {\bf Src}             &  seitdem habe ich sieben H\"{a}user in der Nachbarschaft mit den Lichtern versorgt und sie funktionierenen wirklich gut .  \\
    Iter 1        &  and I 've \underline{been seven homes since in} neighborhood with the lights and they 're really functional . \\
    Iter 2        &  and I 've \underline{been seven homes in the} neighborhood with the lights , and they 're a really functional .  \\
    Iter 4        &  and I 've \underline{been seven homes in} neighborhood with the lights , and they 're a really functional .      \\
    Iter 8        & and I 've \underline{been providing seven homes in the} neighborhood with the lights and they 're a really functional .   \\
    Iter 20		& and I 've \underline{been providing seven homes in the} neighborhood with the lights , and they 're a very good functional . \\
    \textbf{Ref} & since now , I 've set up seven homes around my community , and they 're really working .  \\
    \bottomrule
    \end{tabular}
\end{minipage}

\begin{minipage}{\textwidth}
\small
        \begin{tabular}{ll}
    \midrule
    {\bf Src}             &  er sah sehr gl\"{u}cklich aus , was damals ziemlich ungew\"{o}hnlich war , da ihn die Nachrichten meistens deprimierten .  \\
    Iter 1        &  he looked very happy , which was pretty \underline{unusual the ,}  because  \underline{the news was were usually} depressing . \\
    Iter 2        &  he looked very happy , which was pretty \underline{unusual at the ,}  because  \underline{the news was s} depressing .  \\
    Iter 4        &  he looked very happy , which was pretty \underline{unusual at the ,}  because \underline{news was mostly} depressing .     \\
    Iter 8        & he looked very happy , which was pretty \underline{unusual at the time} because  \underline{the news was mostly} depressing .   \\
    Iter 20		& he looked very happy , which was pretty \underline{unusual at the time ,} because \underline{the news was mostly} depressing . \\
    \textbf{Ref} & there was a big smile on his face which was unusual then , because the news mostly depressed him .  \\
    \bottomrule
    \end{tabular}
\end{minipage}

\begin{minipage}{0.48\textwidth}
\small
    \begin{tabular}{ll}
    \midrule
    {\bf Src}             &  furchtlos zu sein hei{\ss}t f{\"u}r mich , heute ehrlich zu sein .  \\
    Iter 1        &  to be \underline{, for me ,} to be honest today . \\
    Iter 2        &  to be \underline{fearless , me , is} to be honest today .  \\
    Iter 4        &  to be \underline{fearless for me , is} to be honest today .    \\
    Iter 8        & to be \underline{fearless for me , me} to be honest today .   \\
    Iter 20		& to be \underline{fearless for me , is} to be honest today . \\
    \textbf{Ref} & so today , for me , being fearless means being honest .  \\
    \bottomrule
    \end{tabular}
\end{minipage}
\hfill
\begin{minipage}{0.47\textwidth}
    \caption{Three sample De$\rightarrow$En translations from the 
    non-autoregressive sequence model. Source sentences are from the dev set of IWSLT'16. The first iteration corresponds to Decoder 1, and from thereon, Decoder 2 is repeatedly applied.
    Sub-sequences with changes across the refinement steps are underlined.}
    \label{tab:examples_translation}
\end{minipage}
\end{table*}
}

\paragraph{Machine Translation}

In Table~\ref{tab:examples_translation}, we present three sample translations and their iterative refinement steps from the development set of IWSLT'16 (De$\rightarrow$En). As expected, the sequence generated from the first iteration is a rough version of translation and 
is iteratively refined over multiple steps. By inspecting the underlined sub-sequences, we see that each iteration does not monotonically improve the translation, but overall modifies the translation towards the reference sentence. Missing words are added, while unnecessary words are dropped. For instance, see the second example. The second iteration removes the unnecessary ``were'', and the fourth iteration inserts a new word ``mostly''. The phrase ``at the time'' is gradually added one word at a time. 

\paragraph{Image Caption Generation}

Table~\ref{tab:examples_coco} shows two examples of image caption generation. 
We observe that each iteration captures more and more details of the input image. In the first example (left), the bus was described only as a ``yellow bus'' in the first iteration, but the subsequent iterations refine it into ``yellow and black bus''. Similarly, ``road'' is refined into  ``lot''.
We notice this behavior in the second example (right) as well. The first iteration does not specify the place in which ``a woman'' is ``standing on'', which is fixed immediately in the second iteration: ``standing on a tennis court''. In the final and fourth iteration, the proposed model captures the fact that the ``woman'' is ``holding'' a racquet. 

\begin{table*}[t]
\begin{minipage}{0.48\textwidth}
\includegraphics[width=0.4\linewidth]{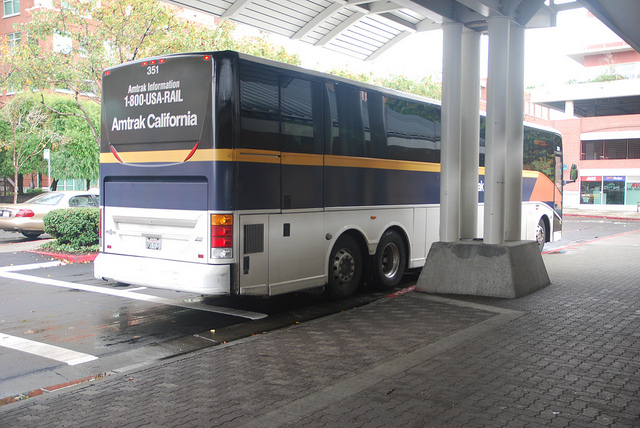}

\centering
\small
    \begin{tabular}{ll}
    \toprule
    \multicolumn{2}{l}{Generated Caption} \\
    \midrule
Iter 1         &  a \underline{yellow bus parked} \underline{on parked in of} parking \underline{road} . \\
Iter 2         &  a \underline{yellow and black} \underline{on parked in a} parking \underline{lot} .   \\
Iter 3         &  a \underline{yellow and black bus} \underline{parked in a} parking \underline{lot} .  \\
Iter 4         & a \underline{yellow and black bus} \underline{parked in a} parking \underline{lot} .   \\
\midrule
    \multicolumn{2}{l}{Reference Captions} \\
    \midrule
\multicolumn{2}{l}{a tour bus is parked on the curb waiting}  \\
\multicolumn{2}{l}{city bus parked on side of hotel in the rain .}  \\
\multicolumn{2}{l}{bus parked under an awning next to brick sidewalk}  \\   
\multicolumn{2}{l}{a bus is parked on the curb in front of a building .}  \\
\multicolumn{2}{l}{a double decked bus sits parked under an awning}  \\
    \bottomrule
    \end{tabular}
\end{minipage}
\begin{minipage}{0.48\textwidth}
\includegraphics[width=0.4\linewidth]{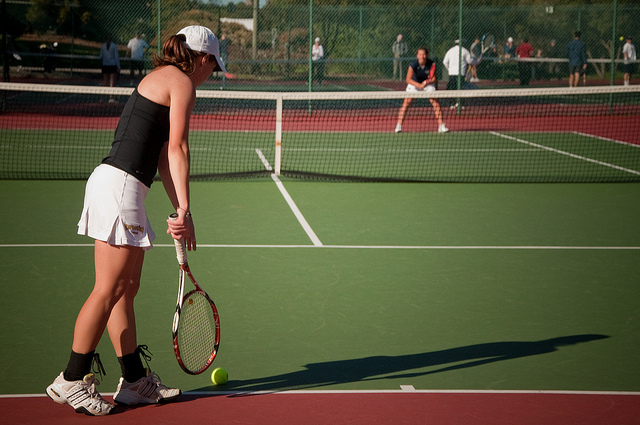}

\centering
\small
    \begin{tabular}{ll}
    \toprule
    \multicolumn{2}{l}{Generated Caption} \\
    \midrule
Iter 1         &  a woman standing on \underline{playing tennis on} a \underline{tennis racquet} . \\
Iter 2         &  a woman standing on \underline{a tennis court} a \underline{tennis racquet} .   \\
Iter 3         &  a woman standing on \underline{a tennis court a} a \underline{racquet} .      \\
Iter 4         & a woman standing on \underline{a tennis court holding} a \underline{racquet} .    \\
\midrule
    \multicolumn{2}{l}{Reference Captions} \\
    \midrule
\multicolumn{2}{l}{a female tennis player in a black top playing tennis}  \\
\multicolumn{2}{l}{a woman standing on a tennis court holding a racquet .}  \\
\multicolumn{2}{l}{a female tennis player preparing to serve the ball .}  \\
\multicolumn{2}{l}{a woman is holding a tennis racket on a court} \\
\multicolumn{2}{l}{a woman getting ready to reach for a tennis ball on the ground} \\
    \bottomrule
    \end{tabular}
\end{minipage}
\vspace{-2mm}
\caption{
Two sample image captions from the proposed non-autoregressive sequence model. The images are from the development set of MS COCO. The first iteration is from decoder 1, while the subsequent ones are from decoder 2. Subsequences with changes across the refinement steps are underlined.
\label{tab:examples_coco}
}
\vspace{-5mm}
\end{table*}

\section{Conclusion}

Following on the exciting, recent success of non-autoregressive neural sequence modeling by \citet{gu2017non} and \citet{oord2017parallel}, we proposed a deterministic non-autoregressive neural sequence model based on the idea of iterative refinement. We designed a learning algorithm specialized to the proposed approach by interpreting the entire model as a latent variable model and each refinement step as denoising. 

We implemented our approach using the 
Transformer
and evaluated it on two tasks: machine translation and image caption generation. On both tasks, we were able to show that the proposed non-autoregressive model performs closely to the autoregressive counterpart with significant speedup in decoding. Qualitative analysis revealed that the 
iterative refinement indeed refines a target sequence gradually over multiple steps. 

Despite these promising results, we observed that proposed non-autoregressive neural sequence model is outperformed by its autoregressive counterpart in terms of the generation quality. 
The following directions should be pursued in the future to narrow this gap. First, we should investigate better approximation to the marginal log-probability.
Second, the impact of the corruption process on the generation quality must be studied. Lastly, further work on sequence-to-sequence model architectures could yield better results in non-autoregressive sequence modeling.

\section*{Acknowledgement}

We thank support by AdeptMind, eBay, TenCent and NVIDIA. This work was partly supported by Samsung Advanced Institute of Technology (Next Generation Deep Learning: from pattern recognition to AI) and Samsung Electronics (Improving Deep Learning using Latent Structure). We also thank Jiatao Gu for valuable feedback.

\balance
\bibliography{main}

\begin{thebibliography}{38}
\expandafter\ifx\csname natexlab\endcsname\relax\def\natexlab#1{#1}\fi

\bibitem[{Alain and Bengio(2014)}]{alain2014regularized}
Guillaume Alain and Yoshua Bengio. 2014.
\newblock What regularized auto-encoders learn from the data-generating
  distribution.
\newblock \emph{The Journal of Machine Learning Research}, 15(1).

\bibitem[{Artetxe et~al.(2017)Artetxe, Labaka, Agirre, and Cho}]{Artetxe:17}
Mikel Artetxe, Gorka Labaka, Eneko Agirre, and Kyunghyun Cho. 2017.
\newblock Unsupervised neural machine translation.
\newblock \emph{arXiv preprint arXiv:1710.11041}.

\bibitem[{Bahdanau et~al.(2014)Bahdanau, Cho, and Bengio}]{bahdanau2014neural}
Dzmitry Bahdanau, Kyunghyun Cho, and Yoshua Bengio. 2014.
\newblock Neural machine translation by jointly learning to align and
  translate.
\newblock \emph{arXiv preprint arXiv:1409.0473}.

\bibitem[{Bordes et~al.(2017)Bordes, Honari, and Vincent}]{pascal_infusion}
Florian Bordes, Sina Honari, and Pascal Vincent. 2017.
\newblock Learning to generate samples from noise through infusion training.
\newblock In \emph{ICLR}.

\bibitem[{Chiu et~al.(2017)Chiu, Sainath, Wu, Prabhavalkar, Nguyen, Chen,
  Kannan, Weiss, Rao, Gonina et~al.}]{chiu2017state}
Chung-Cheng Chiu, Tara~N Sainath, Yonghui Wu, Rohit Prabhavalkar, Patrick
  Nguyen, Zhifeng Chen, Anjuli Kannan, Ron~J Weiss, Kanishka Rao, Katya Gonina,
  et~al. 2017.
\newblock State-of-the-art speech recognition with sequence-to-sequence models.
\newblock \emph{arXiv preprint arXiv:1712.01769}.

\bibitem[{Cho(2016)}]{cho2016noisy}
Kyunghyun Cho. 2016.
\newblock Noisy parallel approximate decoding for conditional recurrent
  language model.
\newblock \emph{arXiv preprint arXiv:1605.03835}.

\bibitem[{Cho et~al.(2015)Cho, Courville, and Bengio}]{cho2015describing}
Kyunghyun Cho, Aaron Courville, and Yoshua Bengio. 2015.
\newblock Describing multimedia content using attention-based encoder-decoder
  networks.
\newblock \emph{IEEE Transactions on Multimedia}, 17(11).

\bibitem[{Chorowski et~al.(2015)Chorowski, Bahdanau, Serdyuk, Cho, and
  Bengio}]{chorowski2015attention}
Jan~K Chorowski, Dzmitry Bahdanau, Dmitriy Serdyuk, Kyunghyun Cho, and Yoshua
  Bengio. 2015.
\newblock Attention-based models for speech recognition.
\newblock In \emph{NIPS}.

\bibitem[{Deng et~al.(2009)Deng, Dong, Socher, Li, Li, and
  Fei-Fei}]{deng2009imagenet}
Jia Deng, Wei Dong, Richard Socher, Li-Jia Li, Kai Li, and Li~Fei-Fei. 2009.
\newblock Imagenet: A large-scale hierarchical image database.
\newblock In \emph{CVPR}.

\bibitem[{Dyer et~al.(2013)Dyer, Chahuneau, and Smith}]{dyer2013simple}
Chris Dyer, Victor Chahuneau, and Noah~A Smith. 2013.
\newblock A simple, fast, and effective reparameterization of ibm model 2.
\newblock In \emph{ACL}.

\bibitem[{Grangier and Auli(2017)}]{quickedit}
David Grangier and Michael Auli. 2017.
\newblock Quickedit: Editing text \& translations via simple delete actions.
\newblock \emph{arXiv preprint arXiv:1711.04805}.

\bibitem[{Gu et~al.(2017)Gu, Bradbury, Xiong, Li, and Socher}]{gu2017non}
Jiatao Gu, James Bradbury, Caiming Xiong, Victor~OK Li, and Richard Socher.
  2017.
\newblock Non-autoregressive neural machine translation.
\newblock \emph{arXiv preprint arXiv:1711.02281}.

\bibitem[{He et~al.(2016)He, Zhang, Ren, and Sun}]{he2016deep}
Kaiming He, Xiangyu Zhang, Shaoqing Ren, and Jian Sun. 2016.
\newblock Deep residual learning for image recognition.
\newblock In \emph{CVPR}.

\bibitem[{Hill et~al.(2016)Hill, Cho, and Korhonen}]{Hill:16}
Felix Hill, Kyunghyun Cho, and Anna Korhonen. 2016.
\newblock Learning distributed representations of sentences from unlabelled
  data.
\newblock In \emph{NAACL}.

\bibitem[{Hinton et~al.(2015)Hinton, Vinyals, and Dean}]{hinton2015distilling}
Geoffrey Hinton, Oriol Vinyals, and Jeff Dean. 2015.
\newblock Distilling the knowledge in a neural network.
\newblock \emph{arXiv preprint arXiv:1503.02531}.

\bibitem[{Hoang et~al.(2017)Hoang, Haffari, and Cohn}]{hoang2017decoding}
Cong Duy~Vu Hoang, Gholamreza Haffari, and Trevor Cohn. 2017.
\newblock Decoding as continuous optimization in neural machine translation.
\newblock \emph{arXiv preprint arXiv:1701.02854}.

\bibitem[{Jouppi et~al.(2017)Jouppi, Young, Patil, Patterson, Agrawal, Bajwa,
  Bates, Bhatia, Boden, Borchers et~al.}]{jouppi2017datacenter}
Norman~P Jouppi, Cliff Young, Nishant Patil, David Patterson, Gaurav Agrawal,
  Raminder Bajwa, Sarah Bates, Suresh Bhatia, Nan Boden, Al~Borchers, et~al.
  2017.
\newblock In-datacenter performance analysis of a tensor processing unit.
\newblock In \emph{Proceedings of the 44th Annual International Symposium on
  Computer Architecture}.

\bibitem[{Kaiser and Bengio(2016)}]{kaiser2016can}
{\L}ukasz Kaiser and Samy Bengio. 2016.
\newblock Can active memory replace attention?
\newblock In \emph{NIPS}.

\bibitem[{Kaiser and Sutskever(2015)}]{kaiser2015neural}
{\L}ukasz Kaiser and Ilya Sutskever. 2015.
\newblock Neural {GPU}s learn algorithms.
\newblock \emph{arXiv preprint arXiv:1511.08228}.

\bibitem[{Kalchbrenner and Blunsom(2013)}]{kalchbrenner2013recurrent}
Nal Kalchbrenner and Phil Blunsom. 2013.
\newblock Recurrent continuous translation models.
\newblock In \emph{EMNLP}.

\bibitem[{Karpathy and Li(2015)}]{karpathy_caption}
Andrej Karpathy and Fei{-}Fei Li. 2015.
\newblock Deep visual-semantic alignments for generating image descriptions.
\newblock In \emph{CVPR}.

\bibitem[{Kim and Rush(2016)}]{kim2016sequence}
Yoon Kim and Alexander~M Rush. 2016.
\newblock Sequence-level knowledge distillation.
\newblock \emph{arXiv preprint arXiv:1606.07947}.

\bibitem[{Kingma and Ba(2014)}]{kingma2014adam}
Diederik~P Kingma and Jimmy Ba. 2014.
\newblock Adam: A method for stochastic optimization.
\newblock \emph{arXiv preprint arXiv:1412.6980}.

\bibitem[{Kingma et~al.(2016)Kingma, Salimans, Jozefowicz, Chen, Sutskever, and
  Welling}]{kingma2016improved}
Diederik~P Kingma, Tim Salimans, Rafal Jozefowicz, Xi~Chen, Ilya Sutskever, and
  Max Welling. 2016.
\newblock Improved variational inference with inverse autoregressive flow.
\newblock In \emph{NIPS}.

\bibitem[{Koehn et~al.(2007)Koehn, Hoang, Birch, Callison-Burch, Federico,
  Bertoldi, Cowan, Shen, Moran, Zens et~al.}]{koehn2007moses}
Philipp Koehn, Hieu Hoang, Alexandra Birch, Chris Callison-Burch, Marcello
  Federico, Nicola Bertoldi, Brooke Cowan, Wade Shen, Christine Moran, Richard
  Zens, et~al. 2007.
\newblock Moses: Open source toolkit for statistical machine translation.
\newblock In \emph{ACL}.

\bibitem[{Lample et~al.(2017)Lample, Denoyer, and Ranzato}]{Lample:17}
Guillaume Lample, Ludovic Denoyer, and Marc'Aurelio Ranzato. 2017.
\newblock Unsupervised machine translation using monolingual corpora only.
\newblock \emph{arXiv preprint arXiv:1711.00043}.

\bibitem[{Lin et~al.(2014)Lin, Maire, Belongie, Hays, Perona, Ramanan,
  Doll{\'a}r, and Zitnick}]{mscoco}
T.Y. Lin, M.~Maire, S.~Belongie, J.~Hays, P.~Perona, D.~Ramanan, P.~Doll{\'a}r,
  and C.~L. Zitnick. 2014.
\newblock Microsoft {COCO}: Common objects in context.
\newblock In \emph{ECCV}.

\bibitem[{Mikolov et~al.(2010)Mikolov, Karafi{\'a}t, Burget,
  {\v{C}}ernock{\`y}, and Khudanpur}]{mikolov2010recurrent}
Tom{\'a}{\v{s}} Mikolov, Martin Karafi{\'a}t, Luk{\'a}{\v{s}} Burget, Jan
  {\v{C}}ernock{\`y}, and Sanjeev Khudanpur. 2010.
\newblock Recurrent neural network based language model.
\newblock In \emph{Eleventh Annual Conference of the International Speech
  Communication Association}.

\bibitem[{Novak et~al.(2016)Novak, Auli, and Grangier}]{auli_iterative_mt}
Roman Novak, Michael Auli, and David Grangier. 2016.
\newblock Iterative refinement for machine translation.
\newblock \emph{arXiv preprint arXiv:1610.06602}.

\bibitem[{Oord et~al.(2016)Oord, Dieleman, Zen, Simonyan, Vinyals, Graves,
  Kalchbrenner, Senior, and Kavukcuoglu}]{van2016wavenet}
Aaron van~den Oord, Sander Dieleman, Heiga Zen, Karen Simonyan, Oriol Vinyals,
  Alex Graves, Nal Kalchbrenner, Andrew Senior, and Koray Kavukcuoglu. 2016.
\newblock Wavenet: A generative model for raw audio.
\newblock \emph{arXiv preprint arXiv:1609.03499}.

\bibitem[{Oord et~al.(2017)Oord, Li, Babuschkin, Simonyan, Vinyals,
  Kavukcuoglu, Driessche, Lockhart, Cobo, Stimberg et~al.}]{oord2017parallel}
Aaron van~den Oord, Yazhe Li, Igor Babuschkin, Karen Simonyan, Oriol Vinyals,
  Koray Kavukcuoglu, George van~den Driessche, Edward Lockhart, Luis~C Cobo,
  Florian Stimberg, et~al. 2017.
\newblock Parallel wavenet: Fast high-fidelity speech synthesis.
\newblock \emph{arXiv preprint arXiv:1711.10433}.

\bibitem[{Papineni et~al.(2002)Papineni, Roukos, Ward, and
  Zhu}]{papineni2002bleu}
Kishore Papineni, Salim Roukos, Todd Ward, and Wei-Jing Zhu. 2002.
\newblock Bleu: a method for automatic evaluation of machine translation.
\newblock In \emph{ACL}.

\bibitem[{Schwenk(2012)}]{schwenk2012continuous}
Holger Schwenk. 2012.
\newblock Continuous space translation models for phrase-based statistical
  machine translation.
\newblock \emph{Proceedings of COLING 2012: Posters}.

\bibitem[{Sennrich et~al.(2016)Sennrich, Haddow, and Birch}]{Sennrich:16}
Rico Sennrich, Barry Haddow, and Alexandra Birch. 2016.
\newblock Neural machine translation of rare words with subword units.
\newblock In \emph{ACL}.

\bibitem[{Srivastava et~al.(2015)Srivastava, Greff, and
  Schmidhuber}]{srivastava2015highway}
Rupesh~Kumar Srivastava, Klaus Greff, and J{\"u}rgen Schmidhuber. 2015.
\newblock Highway networks.
\newblock \emph{arXiv preprint arXiv:1505.00387}.

\bibitem[{Sutskever et~al.(2014)Sutskever, Vinyals, and
  Le}]{sutskever2014sequence}
Ilya Sutskever, Oriol Vinyals, and Quoc~V Le. 2014.
\newblock Sequence to sequence learning with neural networks.
\newblock In \emph{NIPS}.

\bibitem[{Vaswani et~al.(2017)Vaswani, Shazeer, Parmar, Uszkoreit, Jones,
  Gomez, Kaiser, and Polosukhin}]{vaswani2017attention}
Ashish Vaswani, Noam Shazeer, Niki Parmar, Jakob Uszkoreit, Llion Jones,
  Aidan~N Gomez, {\L}ukasz Kaiser, and Illia Polosukhin. 2017.
\newblock Attention is all you need.
\newblock In \emph{NIPS}.

\bibitem[{Xia et~al.(2017)Xia, Tian, Wu, Lin, Qin, Yu, and
  Liu}]{xia2017deliberation}
Yingce Xia, Fei Tian, Lijun Wu, Jianxin Lin, Tao Qin, Nenghai Yu, and Tie-Yan
  Liu. 2017.
\newblock Deliberation networks: Sequence generation beyond one-pass decoding.
\newblock In \emph{NIPS}.

\end{thebibliography}
\bibliographystyle{acl_natbib_nourl}

\appendix

\clearpage

\section{Impact of Length Prediction}
\label{sec:lenpred}

The quality of length prediction has an impact on the overall translation/captioning performance. When using the reference target length (during inference), we consistently observed approximately 1 BLEU score improvement over reported results in the tables and figures across different datasets in the paper (see Table~\ref{tab:length_prediction} for more detailed comparison).

We additionally compared our length prediction model with a simple baseline that uses length statistics of the corresponding training dataset (a non-parametric approach). To predict the target length for a source sentence with length $L_s$, we take the average length of all the target sentences coupled with the sources sentences of length $L_s$ in the training set. Compared to this approach, our length prediction model predicts target length correctly twice as often (16\% vs. 8\%), and gives higher prediction accuracy within five tokens (83\% vs. 69\%)

\begin{table}[h!]
    \centering
    \begin{tabular}[b]{r|rr|rr|rr}
    \toprule
    & \multicolumn{2}{c|}{IWSLT'16} & \multicolumn{2}{c|}{WMT'16} & \multicolumn{2}{c}{WMT'14} \\
    & En$\rightarrow$ & $\rightarrow$En & En$\rightarrow$ & $\rightarrow$En & En$\rightarrow$ & $\rightarrow$En \\
    \midrule
    pred & 27.01 & 32.43 & 29.66 & 30.30 & 21.54 & 25.43 \\
    ref & 28.15 & 33.11 & 30.42 & 31.26 & 22.10 & 26.40 \\
    \bottomrule
	\end{tabular}
	\caption{BLEU scores on each dataset when using reference length (ref) and predicted target length (pred).}
	\label{tab:length_prediction}
\end{table}

\end{document}


\maketitle
\begin{abstract}
  This document contains the instructions for preparing a camera-ready
  manuscript for the proceedings of \confname{}. The document itself
  conforms to its own specifications, and is therefore an example of
  what your manuscript should look like. These instructions should be
  used for both papers submitted for review and for final versions of
  accepted papers.  Authors are asked to conform to all the directions
  reported in this document.
\end{abstract}

\section{Credits}

This document has been adapted from the instructions for earlier ACL
and NAACL proceedings. It represents a recent build from
\url{https://github.com/acl-org/acl-pub}, with modifications by Micha
Elsner and Preethi Raghavan, based on the NAACL 2018 instructions by
Margaret Michell and Stephanie Lukin, 2017/2018 (NA)ACL bibtex
suggestions from Jason Eisner, ACL 2017 by Dan Gildea and Min-Yen Kan,
NAACL 2017 by Margaret Mitchell, ACL 2012 by Maggie Li and Michael
White, those from ACL 2010 by Jing-Shing Chang and Philipp Koehn,
those for ACL 2008 by Johanna D. Moore, Simone Teufel, James Allan,
and Sadaoki Furui, those for ACL 2005 by Hwee Tou Ng and Kemal
Oflazer, those for ACL 2002 by Eugene Charniak and Dekang Lin, and
earlier ACL and EACL formats.  Those versions were written by several
people, including John Chen, Henry S. Thompson and Donald
Walker. Additional elements were taken from the formatting
instructions of the {\em International Joint Conference on Artificial
  Intelligence} and the \emph{Conference on Computer Vision and
  Pattern Recognition}.

\section{Introduction}

The following instructions are directed to authors of papers submitted
to \confname{} or accepted for publication in its proceedings. All
authors are required to adhere to these specifications. Authors are
required to provide a Portable Document Format (PDF) version of their
papers. \textbf{The proceedings are designed for printing on A4
paper.}

\section{General Instructions}

Manuscripts must be in two-column format.  Exceptions to the
two-column format include the title, authors' names and complete
addresses, which must be centered at the top of the first page, and
any full-width figures or tables (see the guidelines in
Subsection~\ref{ssec:first}). {\bf Type single-spaced.}  Start all
pages directly under the top margin. See the guidelines later
regarding formatting the first page.  The manuscript should be
printed single-sided and its length
should not exceed the maximum page limit described in Section~\ref{sec:length}.
Pages are numbered for  initial submission. However, {\bf do not number the pages in the camera-ready version}.

By uncommenting {\small\verb|\aclfinalcopy|} at the top of this 
 document, it will compile to produce an example of the camera-ready formatting; by leaving it commented out, the document will be anonymized for initial submission.

The review process is double-blind, so do not include any author information (names, addresses) when submitting a paper for review.  
However, you should maintain space for names and addresses so that they will fit in the final (accepted) version.  The \confname{} \LaTeX\ style will create a titlebox space of 2.5in for you when {\small\verb|\aclfinalcopy|} is commented out.  

The author list for submissions should include all (and only) individuals who made substantial contributions to the work presented. Each author listed on a submission to \confname{} will be notified of submissions, revisions and the final decision. No authors may be added to or removed from submissions to \confname{} after the submission deadline.

\subsection{The Ruler}
The \confname{} style defines a printed ruler which should be presented in the
version submitted for review.  The ruler is provided in order that
reviewers may comment on particular lines in the paper without
circumlocution.  If you are preparing a document without the provided
style files, please arrange for an equivalent ruler to
appear on the final output pages.  The presence or absence of the ruler
should not change the appearance of any other content on the page.  The
camera ready copy should not contain a ruler. (\LaTeX\ users may uncomment the {\small\verb|\aclfinalcopy|} command in the document preamble.)  

Reviewers: note that the ruler measurements do not align well with
lines in the paper -- this turns out to be very difficult to do well
when the paper contains many figures and equations, and, when done,
looks ugly. In most cases one would expect that the approximate
location will be adequate, although you can also use fractional
references ({\em e.g.}, the first paragraph on this page ends at mark $108.5$).

\subsection{Electronically-available resources}

\conforg{} provides this description in \LaTeX2e{} ({\small\tt
  emnlp2018.tex}) and PDF format ({\small\tt emnlp2018.pdf}), along
with the \LaTeX2e{} style file used to format it ({\small\tt
  emnlp2018.sty}) and an ACL bibliography style ({\small\tt
  acl\_natbib\_nourl.bst}) and example bibliography ({\small\tt
  emnlp2018.bib}).  These files are all available at
\url{http://emnlp2018.org/downloads/emnlp18-latex.zip}; a Microsoft
Word template file ({\small\tt emnlp18-word.docx}) and example
submission pdf ({\small\tt emnlp18-word.pdf}) is available at
\url{http://emnlp2018.org/downloads/emnlp18-word.zip}.  We strongly
recommend the use of these style files, which have been appropriately
tailored for the \confname{} proceedings.

\subsection{Format of Electronic Manuscript}
\label{sect:pdf}

For the production of the electronic manuscript you must use Adobe's
Portable Document Format (PDF). PDF files are usually produced from
\LaTeX\ using the \textit{pdflatex} command. If your version of
\LaTeX\ produces Postscript files, you can convert these into PDF
using \textit{ps2pdf} or \textit{dvipdf}. On Windows, you can also use
Adobe Distiller to generate PDF.

Please make sure that your PDF file includes all the necessary fonts
(especially tree diagrams, symbols, and fonts with Asian
characters). When you print or create the PDF file, there is usually
an option in your printer setup to include none, all or just
non-standard fonts.  Please make sure that you select the option of
including ALL the fonts. \textbf{Before sending it, test your PDF by
  printing it from a computer different from the one where it was
  created.} Moreover, some word processors may generate very large PDF
files, where each page is rendered as an image. Such images may
reproduce poorly. In this case, try alternative ways to obtain the
PDF. One way on some systems is to install a driver for a postscript
printer, send your document to the printer specifying ``Output to a
file'', then convert the file to PDF.

It is of utmost importance to specify the \textbf{A4 format} (21 cm
x 29.7 cm) when formatting the paper. When working with
{\tt dvips}, for instance, one should specify {\tt -t a4}.
Or using the command \verb|\special{papersize=210mm,297mm}| in the latex
preamble (directly below the \verb|\usepackage| commands). Then using 
{\tt dvipdf} and/or {\tt pdflatex} which would make it easier for some.

Print-outs of the PDF file on A4 paper should be identical to the
hardcopy version. If you cannot meet the above requirements about the
production of your electronic submission, please contact the
publication chairs as soon as possible.

\subsection{Layout}
\label{ssec:layout}

Format manuscripts two columns to a page, in the manner these
instructions are formatted. The exact dimensions for a page on A4
paper are:

\begin{itemize}
\item Left and right margins: 2.5 cm
\item Top margin: 2.5 cm
\item Bottom margin: 2.5 cm
\item Column width: 7.7 cm
\item Column height: 24.7 cm
\item Gap between columns: 0.6 cm
\end{itemize}

\noindent Papers should not be submitted on any other paper size.
 If you cannot meet the above requirements about the production of 
 your electronic submission, please contact the publication chairs 
 above as soon as possible.

\subsection{Fonts}

For reasons of uniformity, Adobe's {\bf Times Roman} font should be
used. In \LaTeX2e{} this is accomplished by putting

\begin{quote}
\begin{verbatim}
\usepackage{times}
\usepackage{latexsym}
\end{verbatim}
\end{quote}
in the preamble. If Times Roman is unavailable, use {\bf Computer
  Modern Roman} (\LaTeX2e{}'s default).  Note that the latter is about
  10\% less dense than Adobe's Times Roman font.

\begin{table}[t!]
\begin{center}
\begin{tabular}{|l|rl|}
\hline \bf Type of Text & \bf Font Size & \bf Style \\ \hline
paper title & 15 pt & bold \\
author names & 12 pt & bold \\
author affiliation & 12 pt & \\
the word ``Abstract'' & 12 pt & bold \\
section titles & 12 pt & bold \\
document text & 11 pt  &\\
captions & 10 pt & \\
abstract text & 10 pt & \\
bibliography & 10 pt & \\
footnotes & 9 pt & \\
\hline
\end{tabular}
\end{center}
\caption{\label{font-table} Font guide. }
\end{table}

\subsection{The First Page}
\label{ssec:first}

Center the title, author's name(s) and affiliation(s) across both
columns. Do not use footnotes for affiliations. Do not include the
paper ID number assigned during the submission process. Use the
two-column format only when you begin the abstract.

{\bf Title}: Place the title centered at the top of the first page, in
a 15-point bold font. (For a complete guide to font sizes and styles,
see Table~\ref{font-table}) Long titles should be typed on two lines
without a blank line intervening. Approximately, put the title at 2.5
cm from the top of the page, followed by a blank line, then the
author's names(s), and the affiliation on the following line. Do not
use only initials for given names (middle initials are allowed). Do
not format surnames in all capitals ({\em e.g.}, use ``Mitchell'' not
``MITCHELL'').  Do not format title and section headings in all
capitals as well except for proper names (such as ``BLEU'') that are
conventionally in all capitals.  The affiliation should contain the
author's complete address, and if possible, an electronic mail
address. Start the body of the first page 7.5 cm from the top of the
page.

The title, author names and addresses should be completely identical
to those entered to the electronical paper submission website in order
to maintain the consistency of author information among all
publications of the conference. If they are different, the publication
chairs may resolve the difference without consulting with you; so it
is in your own interest to double-check that the information is
consistent.

{\bf Abstract}: Type the abstract at the beginning of the first
column. The width of the abstract text should be smaller than the
width of the columns for the text in the body of the paper by about
0.6 cm on each side. Center the word {\bf Abstract} in a 12 point bold
font above the body of the abstract. The abstract should be a concise
summary of the general thesis and conclusions of the paper. It should
be no longer than 200 words. The abstract text should be in 10 point font.

{\bf Text}: Begin typing the main body of the text immediately after
the abstract, observing the two-column format as shown in

the present document. Do not include page numbers.

{\bf Indent}: Indent when starting a new paragraph, about 0.4 cm. Use 11 points for text and subsection headings, 12 points for section headings and 15 points for the title.

\begin{table}
\centering
\small
\begin{tabular}{cc}
\begin{tabular}{|l|l|}
\hline
{\bf Command} & {\bf Output}\\\hline
\verb|{\"a}| & {\"a} \\
\verb|{\^e}| & {\^e} \\
\verb|{\`i}| & {\`i} \\ 
\verb|{\.I}| & {\.I} \\ 
\verb|{\o}| & {\o} \\
\verb|{\'u}| & {\'u}  \\ 
\verb|{\aa}| & {\aa}  \\\hline
\end{tabular} & 
\begin{tabular}{|l|l|}
\hline
{\bf Command} & {\bf  Output}\\\hline
\verb|{\c c}| & {\c c} \\ 
\verb|{\u g}| & {\u g} \\ 
\verb|{\l}| & {\l} \\ 
\verb|{\~n}| & {\~n} \\ 
\verb|{\H o}| & {\H o} \\ 
\verb|{\v r}| & {\v r} \\ 
\verb|{\ss}| & {\ss} \\\hline
\end{tabular}
\end{tabular}
\caption{Example commands for accented characters, to be used in, {\em e.g.}, \BibTeX\ names.}\label{tab:accents}
\end{table}

\subsection{Sections}

{\bf Headings}: Type and label section and subsection headings in the
style shown on the present document.  Use numbered sections (Arabic
numerals) in order to facilitate cross references. Number subsections
with the section number and the subsection number separated by a dot,
in Arabic numerals.
Do not number subsubsections.

\begin{table*}[t!]
\centering
\begin{tabular}{lll}
  output & natbib & previous \conforg{} style files\\
  \hline
  \citep{Gusfield:97} & \verb|\citep| & \verb|\cite| \\
  \citet{Gusfield:97} & \verb|\citet| & \verb|\newcite| \\
  \citeyearpar{Gusfield:97} & \verb|\citeyearpar| & \verb|\shortcite| \\
\end{tabular}
\caption{Citation commands supported by the style file.
  The citation style is based on the natbib package and
  supports all natbib citation commands.
  It also supports commands defined in previous \conforg{} style files
  for compatibility.
  }
\end{table*}

{\bf Citations}: Citations within the text appear in parentheses
as~\cite{Gusfield:97} or, if the author's name appears in the text
itself, as Gusfield~\shortcite{Gusfield:97}.
Using the provided \LaTeX\ style, the former is accomplished using
{\small\verb|\cite|} and the latter with {\small\verb|\shortcite|} or {\small\verb|\newcite|}. Collapse multiple citations as in~\cite{Gusfield:97,Aho:72}; this is accomplished with the provided style using commas within the {\small\verb|\cite|} command, {\em e.g.}, {\small\verb|\cite{Gusfield:97,Aho:72}|}. Append lowercase letters to the year in cases of ambiguities.  
 Treat double authors as
in~\cite{Aho:72}, but write as in~\cite{Chandra:81} when more than two
authors are involved. Collapse multiple citations as
in~\cite{Gusfield:97,Aho:72}. Also refrain from using full citations
as sentence constituents.

We suggest that instead of
\begin{quote}
  ``\cite{Gusfield:97} showed that ...''
\end{quote}
you use
\begin{quote}
``Gusfield \shortcite{Gusfield:97}   showed that ...''
\end{quote}

If you are using the provided \LaTeX{} and Bib\TeX{} style files, you
can use the command \verb|\citet| (cite in text)
to get ``author (year)'' citations.

If the Bib\TeX{} file contains DOI fields, the paper
title in the references section will appear as a hyperlink
to the DOI, using the hyperref \LaTeX{} package.
To disable the hyperref package, load the style file
with the \verb|nohyperref| option: \\{\small
\verb|\usepackage[nohyperref]{acl2018}|}

\textbf{Digital Object Identifiers}: As part of our work to make ACL
materials more widely used and cited outside of our discipline, ACL
has registered as a CrossRef member, as a registrant of Digital Object
Identifiers (DOIs), the standard for registering permanent URNs for
referencing scholarly materials. \conforg{} has \textbf{not} adopted the
ACL policy of requiring camera-ready references to contain the appropriate
  DOIs (or as a second resort, the hyperlinked ACL Anthology
  Identifier). But we certainly encourage you to use
  Bib\TeX\ records that contain DOI or URLs for any of the ACL
  materials that you reference. Appropriate records should be found
for most materials in the current ACL Anthology at
\url{http://aclanthology.info/}.

As examples, we cite \cite{P16-1001} to show you how papers with a DOI
will appear in the bibliography.  We cite \cite{C14-1001} to show how
papers without a DOI but with an ACL Anthology Identifier will appear
in the bibliography.  

\textbf{Anonymity:} As reviewing will be double-blind, the submitted
version of the papers should not include the authors' names and
affiliations. Furthermore, self-references that reveal the author's
identity, {\em e.g.},
\begin{quote}
``We previously showed \cite{Gusfield:97} ...''  
\end{quote}
should be avoided. Instead, use citations such as 
\begin{quote}
``\citeauthor{Gusfield:97} \shortcite{Gusfield:97}
previously showed ... ''
\end{quote}

Preprint servers such as arXiv.org and workshops that do not
have published proceedings are not considered archival for purposes of
submission. However, to preserve the spirit of blind review, authors
are encouraged to refrain from posting until the completion of the
review process. Otherwise, authors must state in the online submission
form the name of the workshop or preprint server and title of the
non-archival version. The submitted version should be suitably
anonymized and not contain references to the prior non-archival
version. Reviewers will be told: ``The author(s) have notified us that
there exists a non-archival previous version of this paper with
significantly overlapping text. We have approved submission under
these circumstances, but to preserve the spirit of blind review, the
current submission does not reference the non-archival version.''

\textbf{Please do not use anonymous citations} and do not include
 when submitting your papers. Papers that do not
conform to these requirements may be rejected without review.

\textbf{References}: Gather the full set of references together under
the heading {\bf References}; place the section before any Appendices,
unless they contain references. Arrange the references alphabetically
by first author, rather than by order of occurrence in the text.
By using a .bib file, as in this template, this will be automatically 
handled for you. See the \verb|\bibliography| commands near the end for more.

Provide as complete a citation as possible, using a consistent format,
such as the one for {\em Computational Linguistics\/} or the one in the 
{\em Publication Manual of the American 
Psychological Association\/}~\cite{APA:83}. Use of full names for
authors rather than initials is preferred. A list of abbreviations
for common computer science journals can be found in the ACM 
{\em Computing Reviews\/}~\cite{ACM:83}.

The \LaTeX{} and Bib\TeX{} style files provided roughly fit the
American Psychological Association format, allowing regular citations, 
short citations and multiple citations as described above.  

\begin{itemize}
\item Example citing an arxiv paper: \cite{rasooli-tetrault-2015}. 
\item Example article in journal citation: \cite{Ando2005}.
\item Example article in proceedings, with location: \cite{borsch2011}.
\item Example article in proceedings, without location: \cite{andrew2007scalable}.
\end{itemize}
See corresponding .bib file for further details.

Submissions should accurately reference prior and related work, including code and data. If a piece of prior work appeared in multiple venues, the version that appeared in a refereed, archival venue should be referenced. If multiple versions of a piece of prior work exist, the one used by the authors should be referenced. Authors should not rely on automated citation indices to provide accurate references for prior and related work.

{\bf Appendices}: Appendices, if any, directly follow the text and the
references (but see above).  Letter them in sequence and provide an
informative title: {\bf Appendix A. Title of Appendix}.

\subsection{URLs}

URLs can be typeset using the \verb|\url| command. However, very long
URLs cause a known issue in which the URL highlighting may incorrectly
cross pages or columns in the document. Please check carefully for
URLs too long to appear in the column, which we recommend you break,
shorten or place in footnotes. Be aware that actual URL should appear
in the text in human-readable format; neither internal nor external
hyperlinks will appear in the proceedings.

\subsection{Footnotes}

{\bf Footnotes}: Put footnotes at the bottom of the page and use 9
point font. They may be numbered or referred to by asterisks or other
symbols.\footnote{This is how a footnote should appear.} Footnotes
should be separated from the text by a line.\footnote{Note the line
separating the footnotes from the text.}

\subsection{Graphics}

{\bf Illustrations}: Place figures, tables, and photographs in the
paper near where they are first discussed, rather than at the end, if
possible.  Wide illustrations may run across both columns.  Color
illustrations are discouraged, unless you have verified that  
they will be understandable when printed in black ink.

{\bf Captions}: Provide a caption for every illustration; number each one
sequentially in the form:  ``Figure 1. Caption of the Figure.'' ``Table 1.
Caption of the Table.''  Type the captions of the figures and 
tables below the body, using 11 point text.

\subsection{Accessibility}
\label{ssec:accessibility}

In an effort to accommodate people who are color-blind (as well as those printing
to paper), grayscale readability for all accepted papers will be
encouraged.  Color is not forbidden, but authors should ensure that
tables and figures do not rely solely on color to convey critical
distinctions. A simple criterion: All curves and points in your figures should be clearly distinguishable without color.




\section{Translation of non-English Terms}

It is also advised to supplement non-English characters and terms
with appropriate transliterations and/or translations
since not all readers understand all such characters and terms.
Inline transliteration or translation can be represented in
the order of: original-form transliteration ``translation''.

\section{Length of Submission}
\label{sec:length}

The \confname{} main conference accepts submissions of long papers and
short papers.
 Long papers may consist of up to eight (8) pages of
content plus unlimited pages for references. Upon acceptance, final
versions of long papers will be given one additional page -- up to nine (9)
pages of content plus unlimited pages for references -- so that reviewers' comments
can be taken into account. Short papers may consist of up to four (4)
pages of content, plus unlimited pages for references. Upon
acceptance, short papers will be given five (5) pages in the
proceedings and unlimited pages for references. 

For both long and short papers, all illustrations and tables that are part
of the main text must be accommodated within these page limits, observing
the formatting instructions given in the present document. Supplementary
material in the form of appendices does not count towards the page limit; see appendix A for further information.

However, note that supplementary material should be supplementary
(rather than central) to the paper, and that reviewers may ignore
supplementary material when reviewing the paper (see Appendix
\ref{sec:supplemental}). Papers that do not conform to the specified
length and formatting requirements are subject to be rejected without
review.

Workshop chairs may have different rules for allowed length and
whether supplemental material is welcome. As always, the respective
call for papers is the authoritative source.

\section*{Acknowledgments}

The acknowledgments should go immediately before the references.  Do
not number the acknowledgments section. Do not include this section
when submitting your paper for review. \\

\noindent {\bf Preparing References:} \\

Include your own bib file like this:
{\small\verb|\bibliographystyle{acl_natbib_nourl}|
\verb|\bibliography{emnlp2018}|}

Where \verb|emnlp2018| corresponds to the {\tt emnlp2018.bib} file.
\bibliography{emnlp2018}
\bibliographystyle{acl_natbib_nourl}

\appendix

\section{Supplemental Material}
\label{sec:supplemental}
Each \confname{} submission can be accompanied by a single PDF
appendix, one {\small\tt.tgz} or {\small\tt.zip} appendix containing
software, and one {\small\tt.tgz} or {\small\tt.zip} appendix
containing data.

Submissions may include resources (software and/or data) used in in
the work and described in the paper. Papers that are submitted with
accompanying software and/or data may receive additional credit toward
the overall evaluation score, and the potential impact of the software
and data will be taken into account when making the
acceptance/rejection decisions. Any accompanying software and/or data
should include licenses and documentation of research review as
appropriate.

\confname{} also encourages the submission of supplementary material
to report preprocessing decisions, model parameters, and other details
necessary for the replication of the experiments reported in the
paper. Seemingly small preprocessing decisions can sometimes make a
large difference in performance, so it is crucial to record such
decisions to precisely characterize state-of-the-art methods.

Nonetheless, supplementary material should be supplementary (rather
than central) to the paper. {\bf Submissions that misuse the supplementary 
material may be rejected without review.}
Essentially, supplementary material may include explanations or details
of proofs or derivations that do not fit into the paper, lists of
features or feature templates, sample inputs and outputs for a system,
pseudo-code or source code, and data. (Source code and data should
be separate uploads, rather than part of the paper).

The paper should not rely on the supplementary material: while the paper
may refer to and cite the supplementary material and the supplementary material will be available to the
reviewers, they will not be asked to review the
supplementary material.

Appendices ({\em i.e.} supplementary material in the form of proofs, tables,
or pseudo-code) should be {\bf uploaded as supplementary material} when submitting the paper for review.
Upon acceptance, the appendices come after the references, as shown here. Use
\verb|\appendix| before any appendix section to switch the section
numbering over to letters.